\documentclass[10pt,twocolumn,letterpaper]{article}

\usepackage{iccv}
\usepackage{times}
\usepackage{epsfig}
\usepackage{graphicx}
\usepackage{amsmath}
\usepackage{amssymb}
\usepackage[lined,ruled,linesnumbered]{algorithm2e}
\usepackage{array,multirow,graphicx}
\usepackage{placeins}
\usepackage{tikz}
\usepackage{caption}

\usepackage{rotating}

\newcommand{\fboxI}{\framebox[3.3\height]{\strut $i$}}
\newcommand{\fboxJ}{\framebox[3.3\height]{\strut $j$}}

\DeclareMathOperator*{\argmin}{arg\,min}
\iccvfinalcopy

\def\iccvPaperID{1043}

\ificcvfinal\fi
\begin{document}

\title{Solving Jigsaw Puzzles with Linear Programming}


\author{Rui Yu \hspace{8mm}
 Chris Russell \hspace{8mm}
Lourdes Agapito\\
\parbox{\textwidth}{\centering  University College London}}

\maketitle
\begin{abstract}
   We propose a novel Linear Program (LP) based formulation for
   solving jigsaw puzzles. We formulate jigsaw solving as a set of
   successive global convex relaxations of the standard NP-hard
   formulation, that can describe both jigsaws with pieces of unknown
   position and puzzles of unknown position and orientation.
   The main contribution and strength of our
   approach comes from the LP assembly strategy.  In contrast to
   existing greedy methods, our LP solver exploits all the pairwise
   matches simultaneously, and computes the position of each
   piece/component globally.  The main advantages of our LP approach
   include: \emph{(i)} a reduced sensitivity to local minima compared
   to greedy approaches, since our successive approximations are global and convex
   and \emph{(ii)} an increased robustness to the presence of mismatches in
   the pairwise matches due to the use of a weighted L1 penalty.  To
   demonstrate the effectiveness of our approach, we test our algorithm on
   public jigsaw datasets and show that it outperforms
   state-of-the-art methods.
\end{abstract}
\section{Introduction}
Jigsaw puzzles were first produced as a form of entertainment around
1760s by a British mapmaker and remain one of the most popular puzzles
today. The goal is to reconstruct the original image from a given
collection of pieces. Although introduced as a game, this problem has
also attracted computer scientists. The first computational approach
to solve this problem dates back to 1964~\cite{Freeman:1964}. It was
later shown that this combinatorial problem is
NP-hard~\cite{Altman:1989,Demaine:2007}. Solving jigsaw puzzles
computationally remains a relevant and intriguing problem noted for
its applications to real-world problems. For instance the automatic
reconstruction of fragmented objects in 3D is of great interest in
archaeology or art restoration, where the artefacts recovered from
sites are often fractured. This problem has attracted significant
attention from the computer graphics community, where solutions have
been offered to the automatic restoration of
frescoes~\cite{Castaneda:etal:VAST11} or the reassembly of fractured
objects in 3D~\cite{Huang:etal:Siggraph06}. 

The sustained interest
in this problem has led to significant progress in automatic puzzle
solvers
~\cite{Gallagher:CVPR2012,Sholomon:2013,Sholomon:2014,Son:ECCV2014}.
 An automatic solver typically consists of two components: the
estimation of pairwise compatibility measures between different pieces,
and the assembly strategy. Most research has been dedicated to finding
better assembly strategies and we continue this trend.
Existing approaches
tend to use either greedy strategies or heuristic global solvers,
e.g.\ genetic algorithms.  While greedy methods are known to be
sensitive to local minima, the performance of genetic algorithms
heavily relies on the choice of fitness function and crossover operation.

In common with the majority of works on solving jigsaw puzzles in the
vision community, we focus on the difficult case in which all pieces
are square, so no information about location or orientation is
provided by the pieces, and the jigsaw must be solved using image
information alone.

\begin{figure*}
\begin{centering}
 \includegraphics[width=2.1\columnwidth]{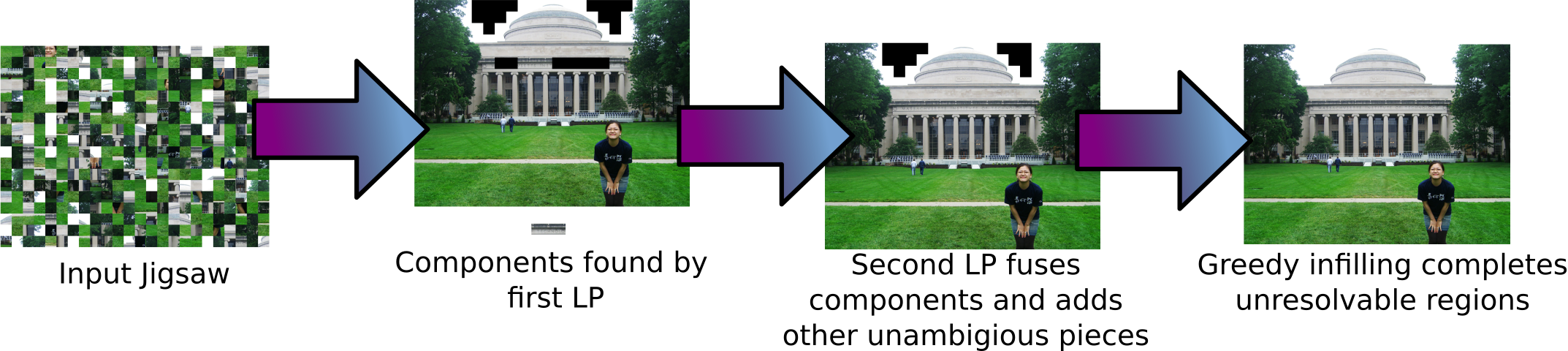}
\end{centering}
\begin{tabular}{p{0.38\columnwidth}p{0.38\columnwidth}p{0.38\columnwidth}p{0.38\columnwidth}p{0.38\columnwidth}}
 \includegraphics[width=0.38\columnwidth]{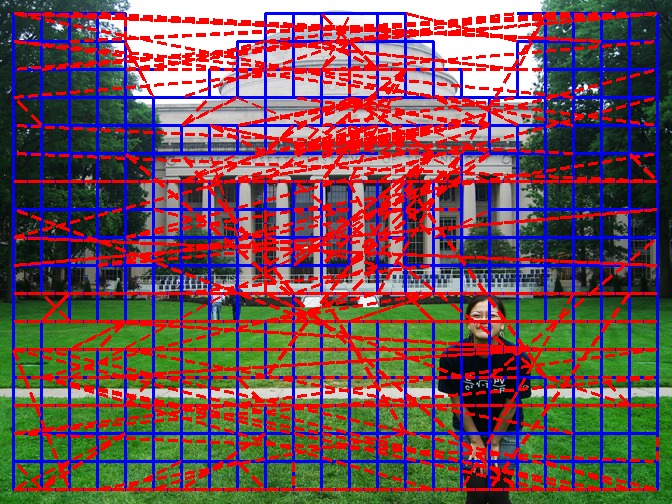}
&
 \includegraphics[width=0.38\columnwidth]{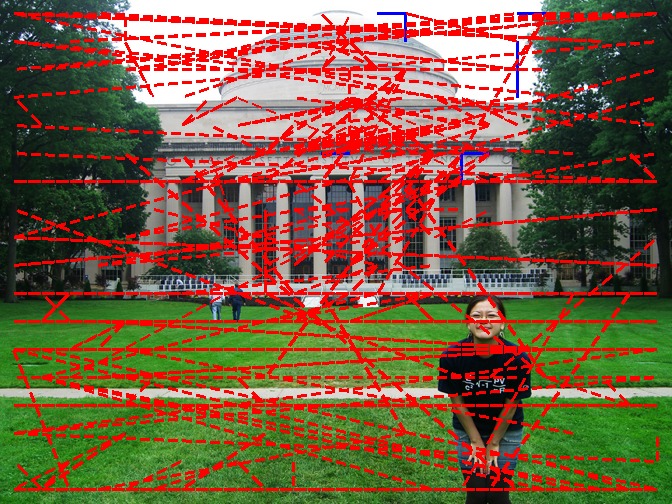}
&
 \includegraphics[width=0.38\columnwidth]{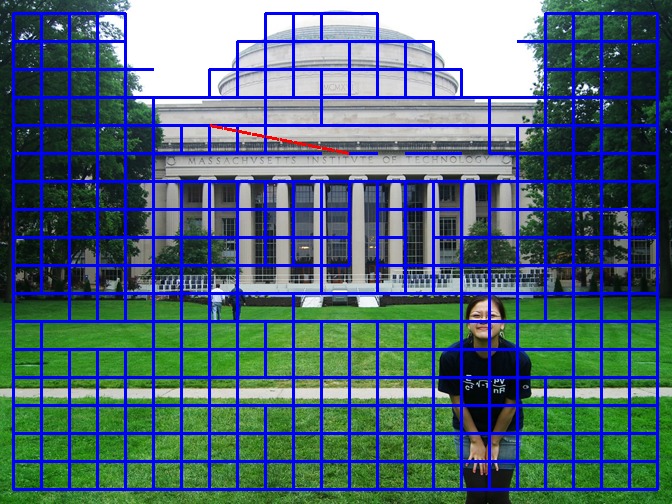}
&
 \includegraphics[width=0.38\columnwidth]{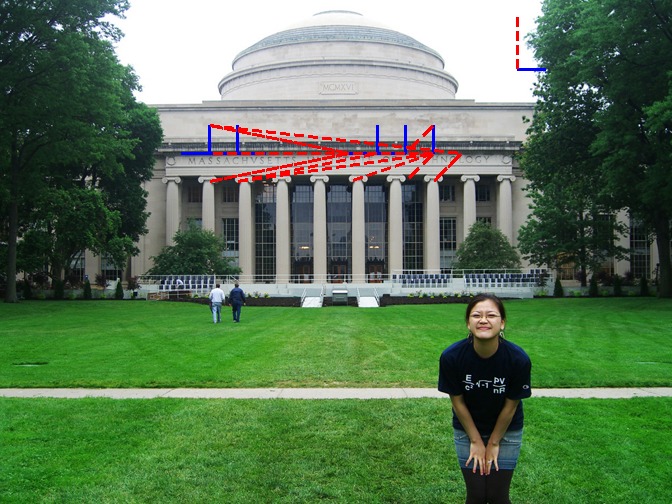}
&
 \includegraphics[width=0.38\columnwidth]{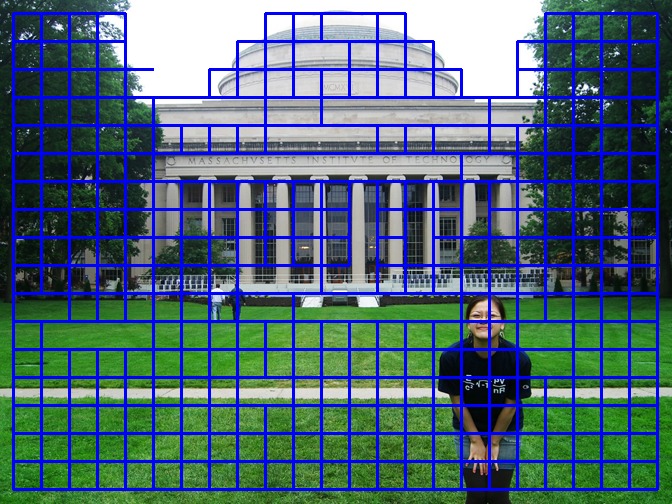}
\\
 
\small Initial confident pairwise matches $A^{(0)}$ &\small Rejected matches $R^{(0)}$ &\small Confident  matches consistent with first LP $A^{(0)}-R^{(0)}$ &\small
 (c) New matches introduced in second LP $A^{(1)}-A^{(0)}$&\small (d) Confident matches consistent with second LP $A^{(1)}-R^{(1)}$\\
\end{tabular}
\caption{Top: An example run through of the algorithm. Bottom: How bad matches are successively eliminated over the same sample run resulting in a better LP. Blue indicates correct ground-truth matches, while red indicates false.  Through-out the run various low confident matches between pieces of sky are present but ignored by the algorithm due to their extremely low weight. For clarity,these are not plotted.   See section~\ref{sec:lp-based-puzzle}.\label{fig:teaser}}
\end{figure*}

In this paper we propose a novel Linear Program (LP) based assembly
strategy. We show that this LP formulation is a convex relaxation of
the original discrete non-convex jigsaw problem. Instead of solving
the difficult NP-hard problem in a single optimization step, we
introduce a sequence of LP relaxations (see figure \ref{fig:teaser}). Our approach proceeds by
alternating between computing new soft constraints that {\em may be
  consistent with the current LP solution} and resolving and
discarding this family of constraints with an LP. Starting with an
initial set of pairwise matches, the method increasingly builds larger
and larger connected components that are consistent with the LP.

Our LP formulation naturally addresses the  so called Type 1
puzzles~\cite{Gallagher:CVPR2012}, where the orientation of each
jigsaw piece is known in advance and only the location of each piece
is unknown. However, we show that our approach
can be directly
extended to the more difficult Type 2 puzzles, where the orientation
of the pieces is also unknown, by first converting it into a Type 1
puzzle with additional pieces.

\section{Related Work}

The first automatic jigsaw solver was introduced by Freeman and
Garder~\cite{Freeman:1964} who proposed a $9$-piece puzzle
solver, that only made use of shape information, and ignored image
data. This was followed by further works~\cite{Wolfson:1988,Webster:1991} that
only made use of shape information.  Kosiba~\cite{Kosiba:1994} was the
first to make use of both shape and appearance information in
evaluating the compatibility between different pieces.

Although it is widely known that jigsaw puzzle problems are NP-hard
when many pairs of pieces are equally likely a priori to be
correct~\cite{Demaine:2007}, many methods take advantage of the
unambiguous nature of visual images to solve jigsaws heuristically.
Existing assembly strategies
can be broadly classified into two main categories:
greedy methods~\cite{Gallagher:CVPR2012,Son:ECCV2014,Yang:2011} and
global methods~\cite{Cho:2010,Andalo:2012,Sholomon:2013,Sholomon:2014}.
The greedy methods start from initial pairwise matches and
successively build larger and larger components,
while global methods directly search for a solution
by maximizing a global compatibility function.

Gallagher~\cite{Gallagher:CVPR2012} proposed a greedy method they
refer to as a \emph{``constrained minimum spanning tree''} based assembly
which greedily merges components while respecting the geometric
consistence constraints. In contrast, Son~\etal~\cite{Son:ECCV2014} showed that exploiting puzzle cycles,
e.g.\ \emph{loop constraints} between the pieces, rather than avoiding
them in tree assembly~\cite{Gallagher:CVPR2012}, leads to better
performance.  They argued that loop constraints could effectively
eliminate pairwise matching outliers and therefore a better solution
could be found by gradually establishing larger consistent puzzle
loops. These loop constraints are related to the loop closure and cycle
preservation constraints used in SLAM~\cite{cummins07icra} and
structure from motion~\cite{Ceylan:etal:TOG13}.
Global approaches include Cho~\etal~\cite{Cho:2010} who
formulated the problem in as a Markov Random Field
and proposed a loopy belief propagation based solution.
Sholomon~\cite{Sholomon:2013,Sholomon:2014} has shown that genetic
algorithms could be successfully applied to large jigsaw puzzles, and
Andalo~\etal\ ~\cite{Andalo:2012} proposed a Quadratic
Programming based jigsaw puzzle solver by relaxing the domain of
permutation matrices to doubly stochastic matrices.

Our approach can be seen as a hybrid between global and greedy
approaches, designed to capture the best of both worlds. Unlike global
approaches which can be hampered by a combinatorial search over the
placement of ambiguous pieces, like greedy methods, we delay the
placing of these pieces until we have information that makes their
position unambiguous. However, unlike greedy methods, at each step of
the algorithm, we consider the placement of all pieces, rather than
just a small initial set of pieces in deciding whether a match is
valid. This makes us less likely to ``paint ourselves into a corner''
with early mistakes. Empirically, the results presented in section
\ref{sec:experiments} reflect this, and we outperform all other
approaches.

\section{Our Puzzle Solver}

In this section we discuss the two principal components of our puzzle 
solver: \emph{(i)} the estimation of pairwise matching costs based on
image intensity information and \emph{(ii)} the assembly strategy. The
fundamental contribution of our work is a novel LP based assembly
algorithm.  After generating initial pairwise matching scores for all
possible pairs of pieces at all four orientations using Gallagher's
approach~\cite{Gallagher:CVPR2012}, our algorithm solves a set of
successive globally convex LP\ relaxations. We initially focus the
description of our algorithm in the case of Type 1 puzzles where only
the position of the pieces is unknown. We will then show how Type 2
puzzles can be solved by converting them into a \emph{mixed} Type 1
puzzle containing copies of each piece in each of the four possible
orientations.

\begin{figure}
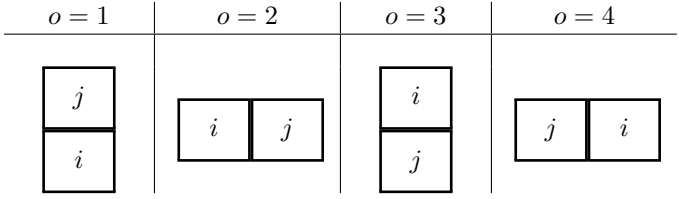

 \centering
 \begin{tabular}{ c| c| c | c }
 $o=1$ &$ o=2$ &$ o=3$ & $o=4$\\
   \hline
   &&&\\
       {
         \begin{tabular}{c}
           \fboxJ
           \\ 
           \fboxI
         \end{tabular}
       }
       &
       {
         \fboxI\fboxJ
       }
       &
       {
         \begin{tabular}{c}
           \fboxI
           \\
           \fboxJ
         \end{tabular}
       }
       &
       {
         \fboxJ\fboxI
       }
       
 \end{tabular}
 \caption{Four different configurations for an oriented pair of matching pieces
   $(i,j,o)$. In this paper we assume that the positive $x$ axis points right
   while the positive $y$ direction points down. From left to right
   the orientation takes values $o=\{1,2,3,4\}$ respectively.} 
 \label{4match}
\end{figure}

\subsection{Matching Constraints}
As shown in Figure~\ref{4match}, given an oriented pair of matching
jigsaw pieces $(i,j,o)$ there are $4$ different possibilities for
their relative orientation $o$. Accordingly, we define $4$
matching constraints on their relative positions on the puzzle grid
$(x_i,y_i)$ and $(x_j,y_j)$, where $x_i$ and $y_i$ indicate the
horizontal and vertical positions of piece $i$. We define the desired
offsets $\delta^x_{o}$, $\delta^y_{o}$ between two pieces in $x$ and $y$ for
an oriented pair $(i,j,o)$ as
  \begin{equation}
    \delta^x_o = 
    \begin{cases}
      0  &  \quad \text{if $o=1$}\\
      -1 &  \quad \text{if $o=2$}\\
      0  &  \quad \text{if $o=3$}\\
      1  &  \quad \text{if $o=4$}\\
    \end{cases} \nonumber
  \end{equation}
  \begin{equation}
    \delta^y_o = 
    \begin{cases}
      1  &  \quad \text{if $o=1$}\\
      0  &  \quad \text{if $o=2$}\\
      -1 & \quad \text{if $o=3$}\\
      0  &  \quad \text{if $o=4$}\\
    \end{cases} \nonumber
  \end{equation}


\subsection{Pairwise Matching Costs}
 To measure the visual compatibility between pairs of jigsaw pieces,
 we use the Mahalanobis Gradient Compatibility (MGC) distance proposed
 by Gallagher~\cite{Gallagher:CVPR2012} which penalizes strong changes
 in image intensity gradients across piece boundaries by encouraging
 similar gradient distributions on either side. This pairwise matching
 cost is widely used by other authors, and therefore allows for a
 direct comparison of our assembly strategy against competing
 approaches~\cite{Gallagher:CVPR2012,Son:ECCV2014}. We define
 $D_{ijo}$ to be the MGC distance (see~\cite{Gallagher:CVPR2012} for
 full definition) between pieces $i$ and $j$ with orientation $o$ (see
 figure~\ref{4match}). The matching weight $w_{ijo}$
 associated with the oriented pair $(i,j,o)$ can now be
 computed as 
\begin{equation}
w_{ijo}=\frac{\min(\min_{k\neq i}(D_{kjo}),\min_{k \neq j}(D_{iko}))}{D_{ijo}},
\end{equation}
i.e. the inverse ratio between $D_{ijo}$ and the best alternative
match. Note that these weights are large when the matching distance
between pieces is relatively small and vice-versa.

\subsection{NP hard objective}
\begin{algorithm}[tb]
\caption{LP based Type 1 jigsaw puzzle solver\label{algo:jigsaw_solver}}
\SetKwInOut{Input}{Input}\SetKwInOut{Output}{Output}
\SetKwInput{Initialisation}{Initialisation}
\SetKwFunction{Median}{Median}
\Input{Scrambled jigsaw puzzle pieces}
\Output{Assembled image}
\Initialisation{Generate initial pairwise matches $U^{(0)}$}
Compute $A^{(0)}$ according to eq~(\ref{eq:Ak})\\
Solve~\eqref{minxLP} to get the initial solution ${\bf x}^{(0)}$, ${\bf y}^{(0)}$\\
\BlankLine
\While{ not converged}
{
  Generate Rejected Matches $R^{(k)}$ using eq~(\ref{eq:Rk})\\
  Update $U^{(k)}$ by discarding $R^{(k)}$\\
  Compute pairwise matches $A^{(k)}$ according to eq~(\ref{eq:Ak})\\
  Solve~\eqref{minxLP}, get new solution ${\bf x}^{(k)}, {\bf y}^{(k)}$ \\
}
Trim and fill as necessary to get rectangular shape~\cite{Gallagher:CVPR2012}
\end{algorithm}

Given the entire set of puzzle pieces $V=\{i \mid i=1,\hdots,N\times M \}$, where $M$ and $N$ are the dimensions of the puzzle, we
define the set 
$U=\{(i,j,o),\,  \forall i\in V,\, \forall j\in V,\, \forall o \in \{0,1,2,3\}\}$ as the universe of all possible oriented matches
$(i,j,o)$ between pieces (see figure~\eqref{4match}). Each oriented match
$(i,j,o)$ will have an associated confidence weight $w_{ijo}$.

We now define the following weighted $L_0$ costs
\begin{equation}\label{jigsaw:energyX}
C ({\bf x})=\sum_{(i,j,o) \in U}w_{ijo}|x_i-x_j-\delta^x_{o}|_0,
\end{equation}
\begin{equation}\label{jigsaw:energyY}
C ({\bf y})=\sum_{(i,j,o) \in U}w_{ijo}|y_i-y_j-\delta^y_{o}|_0,
\end{equation}
%
and formulate the jigsaw puzzle problem as
\begin{align}
{\text{minimize: }}&C ({\bf x}) + C ({\bf y}) \label{jigsaw_cost} \\
 \text{subject to: }
& \forall i, x_i \in \mathbb{N}, 1 \leq x_i \leq M \label{box:const1} \\
& \forall j, y_j \in \mathbb{N}, 1 \leq y_j \leq N \label{box:const2}\\
& \forall i, \forall j, \left| x_i-x_j \right| + \left| y_i-y_j \right| > 0 \label{collision}
\end{align}
where $x_{i}$ and
$y_{i}$ are the $x$ and $y$ position of piece $i$ and 
${\bf x}=[x_1,x_2,\ldots x_{n}]$, ${\bf y}=[y_1,y_2,\ldots y_{n}]$,
with $n$ being the total number of pieces.

The objective function~\eqref{jigsaw_cost} is the sum of cost over $x$ and $y$ dimensions.
The first two constraints~(\ref{box:const1},~\ref{box:const2}) are integer position constraints, that force each
piece to lie on an integer solution inside the puzzle domain. The final constraint~\eqref{collision} guarantees that no two pieces are located in the same position.
Problem~\eqref{jigsaw_cost} can be explained as searching for a configuration
of all the pieces which gives the minimum weighted number of incorrect 
matches. However, both the feasible region and the objective~\eqref{jigsaw_cost}
are discrete and non-convex, and the problem is NP-hard.

\begin{table*}[t]
\caption{\label{table:Type1OlmosPomeranz} Reconstruction performance on Type 1 puzzles from the Olmos\cite{Olmos:2003} and Pomeranz\cite{Pomeranz:2011} datasets. The size of each piece is P = 28 pixels. See Section~\ref{sec:experiments} for a definition of the error measures.}
\begin{center}
\begin{tabular}{|c||c|c||c|c||c|c||c|c|}
\hline
& \multicolumn{2}{c||}{540 pieces} & \multicolumn{2}{c||}{805 pieces} & \multicolumn{2}{c||}{2360 pieces} & \multicolumn{2}{c|}{3300 pieces} \\ \cline{2-9}
& Direct & Nei. & Direct & Nei. & Direct & Nei. & Direct & Nei. \\ \hline
Pomeranz~\etal~\cite{Pomeranz:2011}&83\% &91\% &80\% &90\% &33\% &84\% &80.7\% &95.0\%\\\hline
Andalo~\etal~\cite{Andalo:2012}&90.6\% &95.3\% &82.5\% &93.4\%&-&-&-&-\\ \hline
Sholomon~\etal~\cite{Sholomon:2013} &90.6\% &94.7\% &92.8\% &95.4\% &82.7\% &87.5\% &65.4\% &91.9\%\\ \hline
 Son~\etal~\cite{Son:ECCV2014} & 92.2\% & 95.2\% & 93.1\% & 94.9\% & 94.4\% & 96.4\% & 92.0\% & 96.4\% \\ \hline\hline
  Constrained & 94.0\% & 96.8\% & 95.4\% & 96.6\% & 94.9\% & 97.6\% & 94.1\% & 97.5\% \\ \hline
 Free & 94.6\% & 97.3\% & 94.2\% & 96.4\% & 94.9\% & 97.6\% & 93.9\% & 97.1\% \\ \hline
 Hybrid &\bf 94.8\% &\bf 97.3\% &\bf 95.2\% &\bf 96.6\% &\bf 94.9\% &\bf 97.6\% &\bf 94.1\% &\bf 97.5\% \\ \hline
\end{tabular}
\end{center}
\end{table*}
\begin{table*}[t]
\caption{\label{table:Type2OlmosPomeranz} Reconstruction performance on Type 2 puzzles from the Olmos and Pomeranz dataset. Each piece is a 28 pixel square.  Note that our slightly worse performance than Son~\etal in 3300 piece jigsaws comes from poor resolution of a patch of sky in a single image. See Section~\ref{sec:experiments} for a definition of the error measures.}
\begin{center}
\begin{tabular}{|c||c|c|c|c|c|c|c|c|}
\hline
& \multicolumn{2}{c|}{540 pieces} & \multicolumn{2}{c|}{805 pieces} & \multicolumn{2}{c|}{2360 pieces} & \multicolumn{2}{c|}{3300 pieces} \\ \hline
& Direct & Nei. & Direct & Nei. & Direct & Nei. & Direct & Nei. \\ \hline
Son~\etal~\cite{Son:ECCV2014} & 89.1\% & 92.5\% & 86.4\% & 88.8\% & 94.0\% & 95.3\% & \bf 89.9\% & \bf 93.4\% \\ \hline \hline
Constrained     & 92.6\%      & 93.1\%      & 91.2\%     & 92.0\%      & 94.4\%     & 95.5\% & 88.4\% & 89.2\% \\ \hline
Free  & 88.0\%      & 89.1\%      & 91.4\%     & 92.5\%      & 94.4\%     & 94.8\% &  89.7\% & 90.2\%  \\ \hline
Hybrid   & \bf  92.8\% & \bf  93.3\% & \bf 91.7\% & \bf  92.9\% & \bf 94.4\% & \bf 95.5\% &  89.7\% &  90.2\% \\ \hline
\end{tabular}
\end{center}
\end{table*}
\subsection{LP based Puzzle Assembly}
\label{sec:lp-based-puzzle}
Figure~\ref{fig:teaser} and Algorithm \ref{algo:jigsaw_solver} gives overviews of our approach.
The method takes scrambled pieces as input and returns an assembled
image. Instead of solving the difficult puzzle problem in a single
optimization step, we propose to solve a sequence of LP convex
relaxations. Our approach is an incremental strategy: starting with
initial pairwise matches, we successively build larger connected
components of puzzle pieces by iterating between LP optimization and
generating new matches by deleting those that are inconsistent with
the current LP solution and proposing new candidates not yet shown to be inconsistent.

In comparison to previous greedy
methods~\cite{Gallagher:CVPR2012,Son:ECCV2014}, our approach has two
main advantages. The first one is that instead of gradually building
larger components in a greedy way, our LP step exploits all the
pairwise matches simultaneously. The second advantage is that due to
the weighted $L_1$ cost function, LP is robust to incorrect pairwise
matches. Son~\etal~\cite{Son:ECCV2014} has shown that
verification of {\em loop constraints} significantly improves the
precision of pairwise matches, and our work can be considered as a
natural generalisation of these \emph{loop constraints} that allows loops of all
shapes and sizes, not necessarily rectangular, to be solved in a single
LP optimization. 
The rest of this section describes in detail the two main components
of our formulation: \emph{(i)} the initial LP formulation for the
scrambled input pieces, and \emph{(ii)} the successive LPs for the
connected components. See Figure~\ref{fig:teaser} for an illustration of the whole process on an image taken from the MIT dataset.

\subsubsection{LP Formulation for Scrambled Pieces}
To make the problem convex, we discard the integer position constraints~(\ref{box:const1},\ref{box:const2}) and
the non-collision constraints~(\ref{collision}) and relax the non-convex $L_0$ norm
objective to its tightest convex relaxation $L_1$ norm.
This gives the objective
\begin{align}\label{energyLP_U}
C^* ({\bf x}, {\bf y}) = & C^* ({\bf x}) + C^* ({\bf y}) \\ =
&\sum_{(i,j,o) \in U}w_{ijo}|x_i-x_j-\delta^x_{o}|_1 \\ +&
\sum_{(i,j,o) \in U}w_{ijo}|y_i-y_j-\delta^y_{o}|_1
\end{align}
In practice, this relaxation does not give rise to a plausible solution as the $L_1$ norm acts in a similar way to the weighted median, and results in a compromised solution that overly favours weaker matches.
Instead we consider the same objective over an active subset of candidate matches $A\subseteq U$.
\begin{align}\label{energyLP}
C^+ ({\bf x}, {\bf y}) = & C^+ ({\bf x}) + C^+ ({\bf y}) \\ =
&\sum_{(i,j,o) \in A}w_{ijo}|x_i-x_j-\delta^x_{o}|_1 \\ +&
\sum_{(i,j,o) \in A}w_{ijo}|y_i-y_j-\delta^y_{o}|_1
\end{align}
Note that if $A$ was chosen as the set of ground-truth neighbors,
solving this LP would return the solution to the jigsaw, and so
finding the optimal set $A$ is N.P. hard. Instead we propose a
sequence of estimators of $A$, which we write as $A^{(k)}$.  In
practice, throughout the LP formulation, we consider two sets of
matches $U^{(k)}$ and $A^{(k)}$ for each LP iteration $k$. The first
set $U^{(k)}$ begins as the universal set of all possible candidate
matches for all orientations and choices of piece $U^{(0)}=U$ and is
of size $4 n^2$. In subsequent iterations of our algorithm, its
$k^\text{th}$ form $U^{(k)}$ steadily decreases in size as our
algorithm runs and rejects matches. The second set of matches
$A^{(k)}$ is the {\em active selection} at iteration $k$ of best
candidate matches taken from $U^{(k)}$. For all iterations, $A^{(k)}$
is always of size $4 n$ and can be computed as
\begin{equation}
\label{eq:Ak}
A^{(k)}=\{(i,j,o)\in U^{(k)} : j=\argmin_{j:(i,j,o)\in U^{(k)}} D_{ijo} \},
\end{equation}
We take advantage of the symmetry between $C^+ ({\bf x})$ and $C^+ ({\bf y})$, and 
 only illustrate in the paper how to solve the $x$ coordinate subproblem.

The minimiser of $C^+ ({\bf x})$ can be written as
\begin{equation}
{\argmin_{\bf x}}
\sum_{(i,j,o) \in A^{(k)}}w_{ijo}|x_i-x_j-\delta^x_{o}| _1.
\label{minx}
\end{equation}
By introducing auxiliary variables ${\bf h}$, we transform the
minimisation of \eqref{minx} into the following linear program
\begin{align}
&{\argmin_{{\bf x},{\bf h}}}
& & \sum_{(i,j,o) \in A^{(k)}}w_{ijo}h_{ijo} \label{minxLP}\\
& \text{subject to}
& & h_{ijo} \geq x_i - x_j - \delta^x_{o}, &(i,j,o) \in A^{(k)}\nonumber\\
& & & h_{ijo} \geq -x_i + x_j + \delta^x_{o}, &(i,j,o) \in A^{(k)}.\nonumber
\end{align}
Given the LP solution, we can recover the image based on the relative
position of all the pieces. However, as the linear program does not
enforce either integer location or collision constraints, pieces may
not be grid aligned and the solution may also contain holes (no pieces
in the corresponding position) or collisions (two or more pieces
assigned to the same position). In practice, unlike most other convex
regularizers such as $L_2^2$, the $L_1$ norm encourages a {\em `winner
  takes all'} solution, and while the absolute location of each piece
is arbitrary, the relative locations between adjacent pieces are very
close to integers ($<10e{-6}$). Holes and collisions between pieces
mostly happen where sufficiently ambiguous matches lead to fully
disconnected regions, or contradictory cues lead to a region being
pulled apart, with the weakest matches broken.

Generally, solving this LP does not give rise to a complete and fully
aligned jigsaw. Instead we build on the solution found and by
iteratively solving a sequence of LP optimizations gradually create
larger and larger recovered components. As the LP robustly detects and
eliminates mismatches, the matches consistent with the current LP
solution can be used to establish accurate connected components of
jigsaw pieces.

\subsubsection{ Successive LPs and Removal of Mismatches}
After the $k^{th}$ LP iteration, we validate the solution and remove
matches from the set $A^{(k)}$ that are inconsistent with the solution
to the LP 
\begin{equation}
\label{eq:Rk}
 R^{(k)}= \{ \forall (i,j,o) \in A^{(k)} :|x_i-x_j-\delta^x_{o}|> 10^{-5} \}.
\end{equation}
We set 
\begin{equation}
\label{eq:U-update}
U^{(k)}=U^{({k-1})}-R^{(k)},
\end{equation}
and estimate $A^{(k)}$ according to equation~\eqref{eq:Ak}.
This procedure is guaranteed to converge. Matches are only ever removed from $U^{(k)}$, and as $U^{(0)}$ is finite, the procedure must eventually stop. In practice, we never observed the algorithm take more than 5 iterations, and typically it converges after two iterations.
\subsubsection{Completing the solution}
Much like other methods~\cite{Gallagher:CVPR2012,Son:ECCV2014}, the solution found is not guaranteed to be rectangular, and may still have some pieces we have been unable to assign. As such, we make use of the same post-processing of Gallagher\cite{Gallagher:CVPR2012}, to generate a complete rectangular solution.
\subsection{Constrained variants of our approach}
In practice, one limitation of the above method is that the updating of $A^{(k)}$ can tear apart previously found connected components. In some cases this is good as it allows for the recovery from previously made errors but it is roughly as likely to break good components as bad.

As such, we evaluate a constrained variant of our approach, that enforces as a hard constraint that all pieces lying in a connected components\footnote{ In very rare cases,
there may be a collision within a single connected component,
we break those matches associated with conflicting pieces and use
the remaining matches to rebuild the connected component.
} must remain in the same relative position to other pieces in the component.
In section~\ref{sec:experiments}, we evaluate these two methods and another that selects the best solution found by these two methods based on how well it minimizes the original cost $C({\bf x})+ C({\bf y})$. A comparison between all three strategies can be seen in the tables of section \ref{sec:experiments}, where we refer to the three approaches as {\em free}, {\em constrained}, or {\em hybrid} respectively.
\begin{figure*}
\begin{center}
\begin{tabular}{cccc}
 \includegraphics[width=0.48\columnwidth]{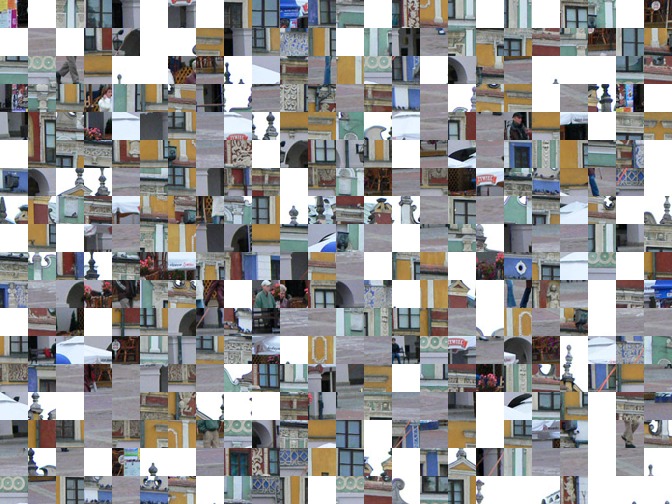}
 &
 \includegraphics[width=0.48\columnwidth]{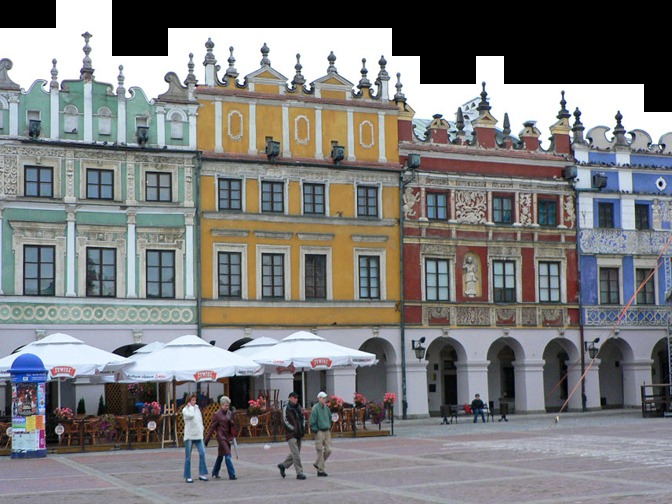}
 &
 \includegraphics[width=0.48\columnwidth]{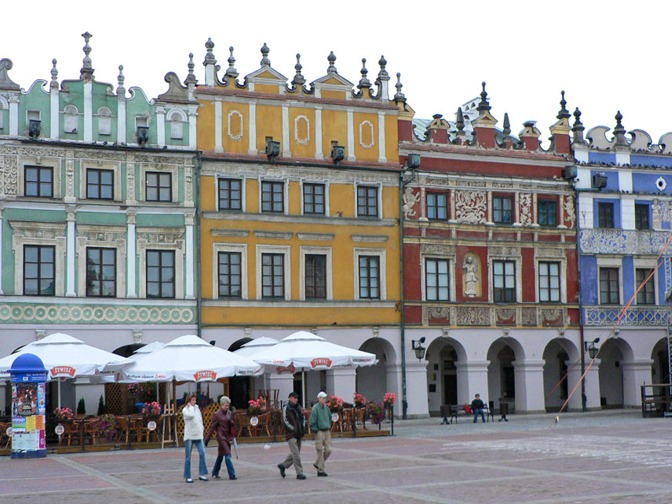}
 &
 \includegraphics[width=0.48\columnwidth]{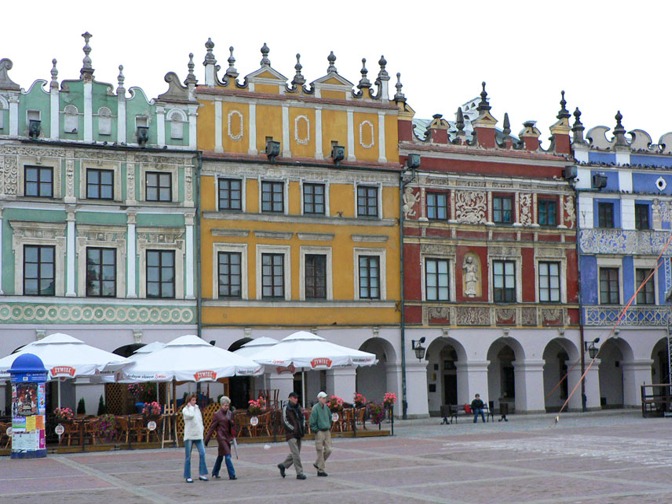}
 \\
 \includegraphics[width=0.48\columnwidth]{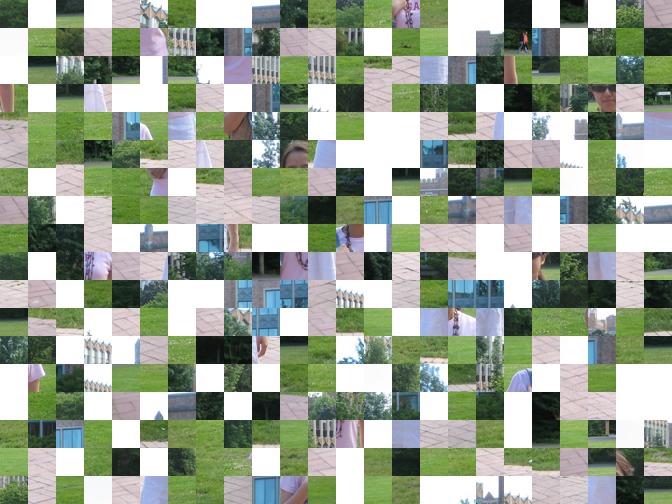}
&
 \includegraphics[width=0.48\columnwidth]{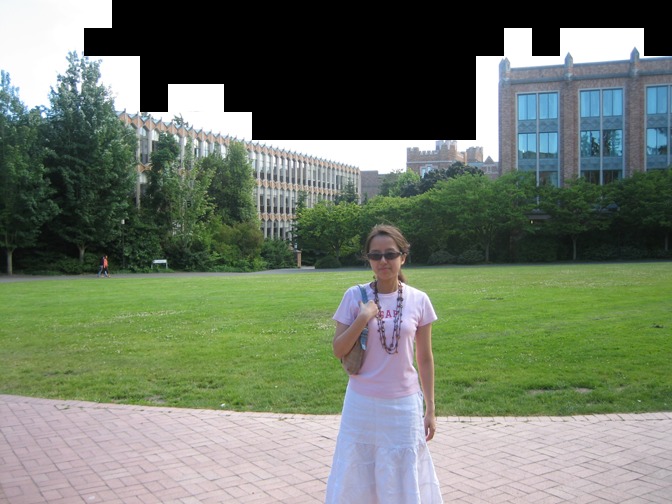}
&
 \includegraphics[width=0.48\columnwidth]{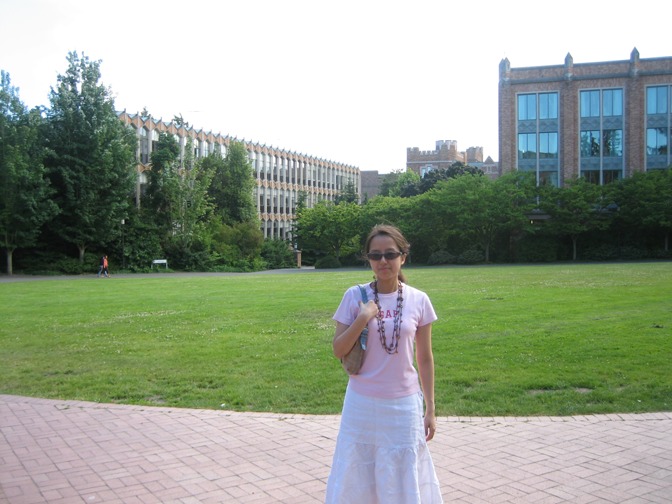}
&
 \includegraphics[width=0.48\columnwidth]{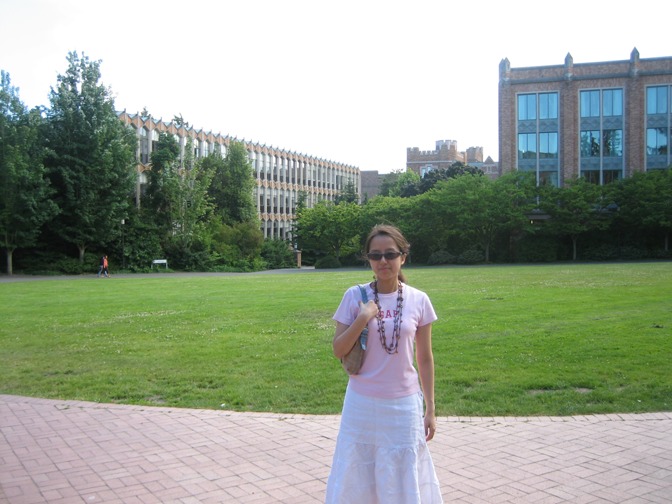}
\\
 (a) & (b) & (c) & (d)
\end{tabular}
\end{center}
\caption{ \label{fig:Fill} Reconstruction results from the initial LP and our complete method,
 images from the MIT dataset. (a) the input pieces, (b) the largest connected
 component before filling, (c) the final result and (d) ground truth.}
\end{figure*}
\begin{figure*}
\begin{center}
\begin{tabular}{ccccc}
  \includegraphics[width=0.36\columnwidth]{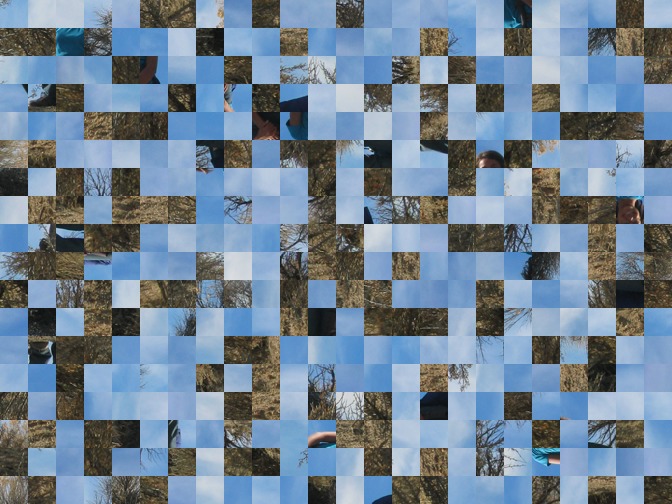}
&
 \includegraphics[height=0.36\columnwidth]{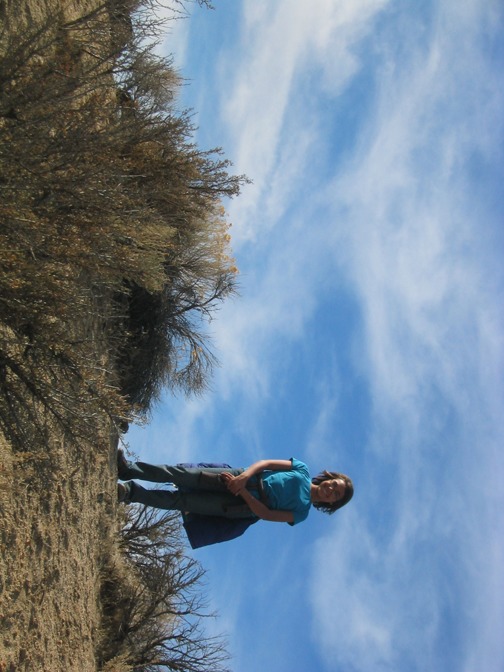}
&
  \includegraphics[width=0.36\columnwidth]{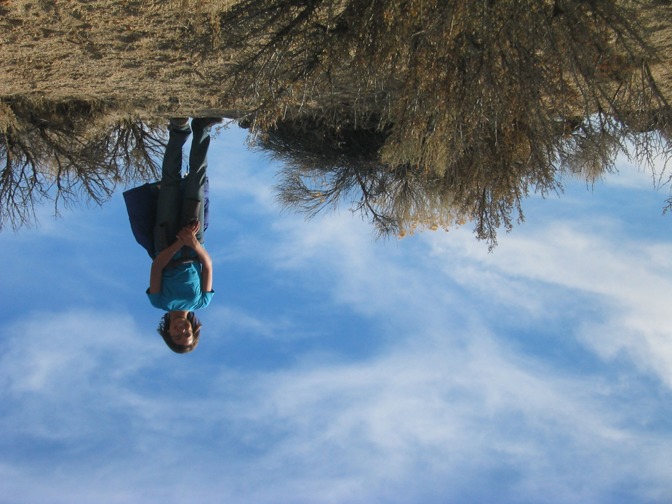}
&
 \includegraphics[height=0.36\columnwidth]{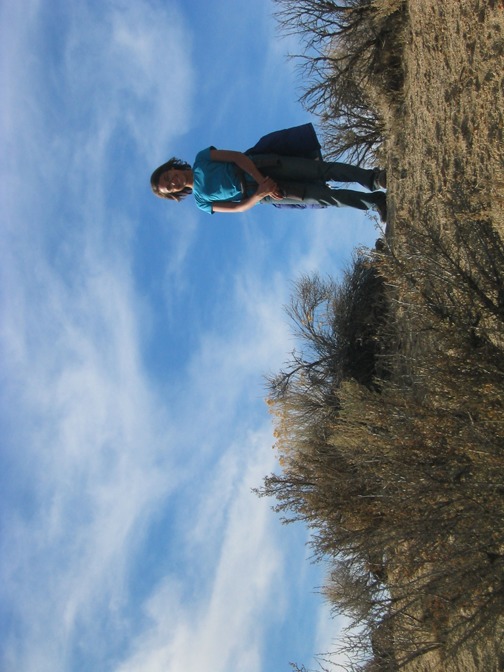}
&
  \includegraphics[width=0.36\columnwidth]{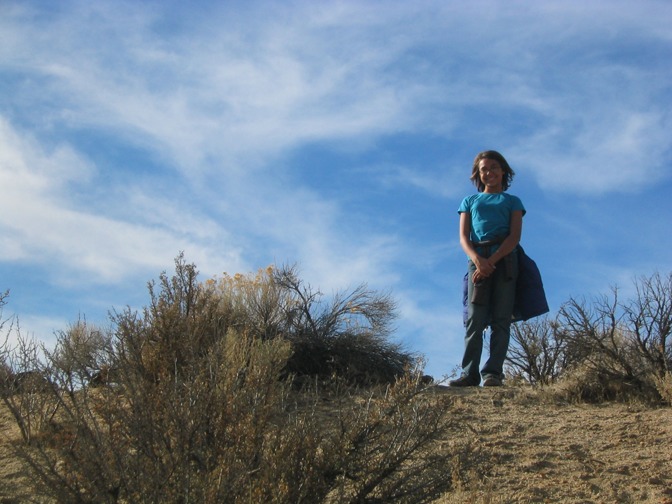}
\end{tabular}
\end{center}
\caption{ \label{fig:Type2} Example reconstruction results of Type 2 Puzzle. From left to right:
the rotated input puzzle pieces, and four recovered images generated by converting to a mixed Type 1 puzzle.}
\end{figure*}
\begin{figure*}
\begin{center}
\begin{tabular}{ccccc}
  \includegraphics[width=0.6\columnwidth,height=0.3\columnwidth]{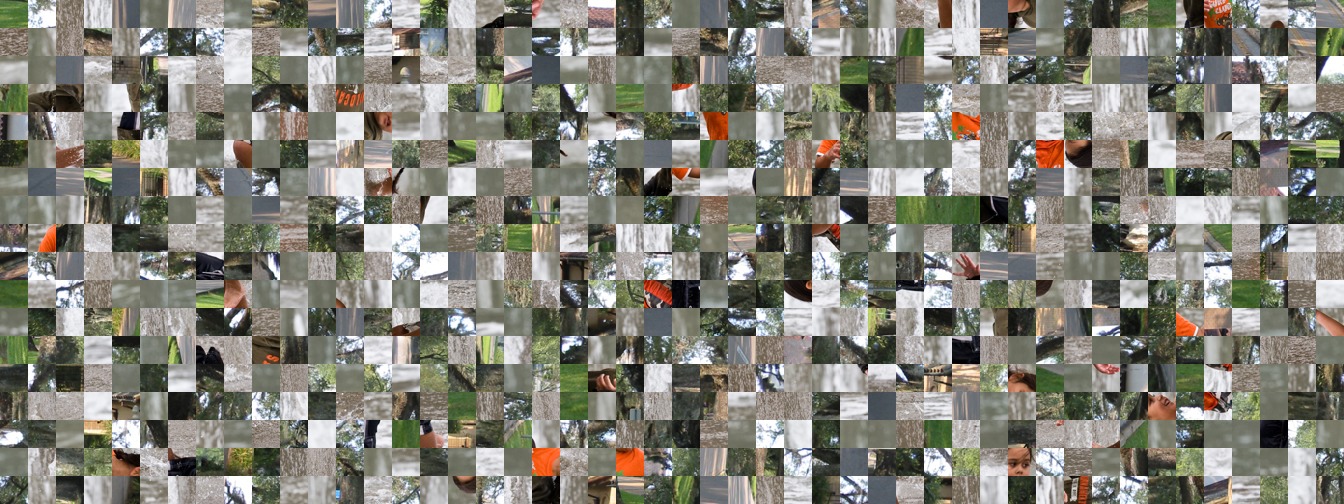}
&
 \includegraphics[width=0.32\columnwidth,height=0.3\columnwidth]{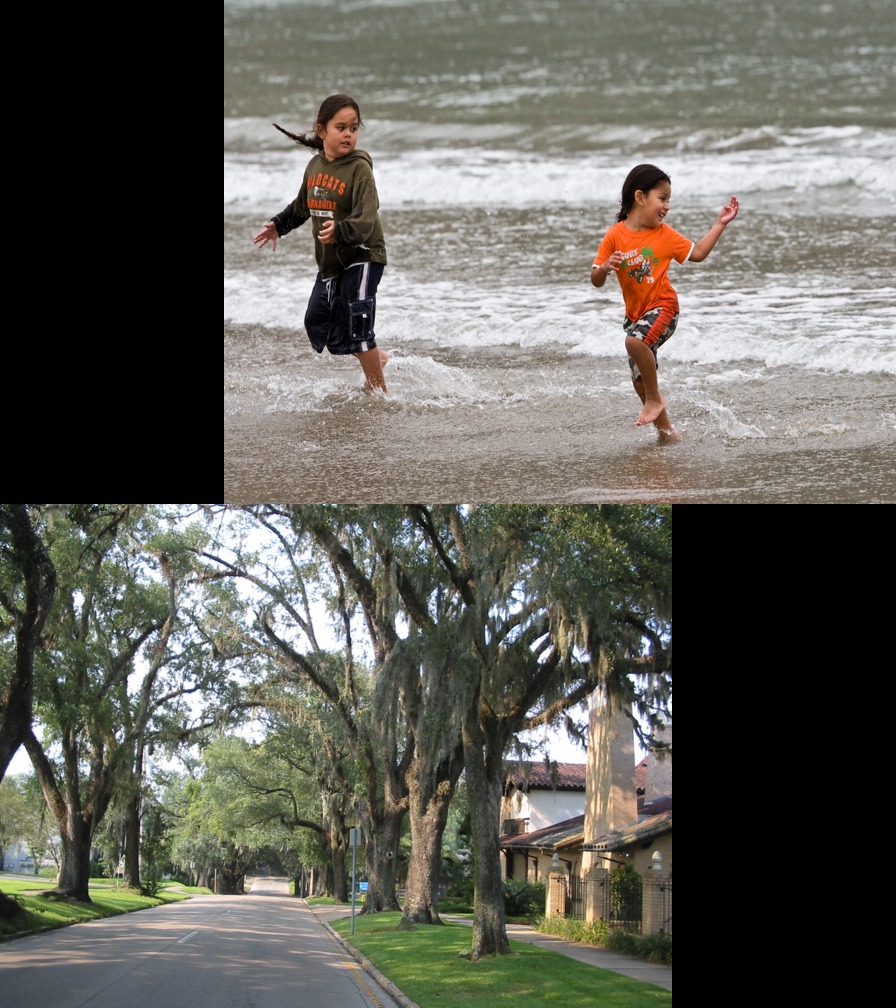}
&
 \includegraphics[width=0.32\columnwidth,height=0.3\columnwidth]{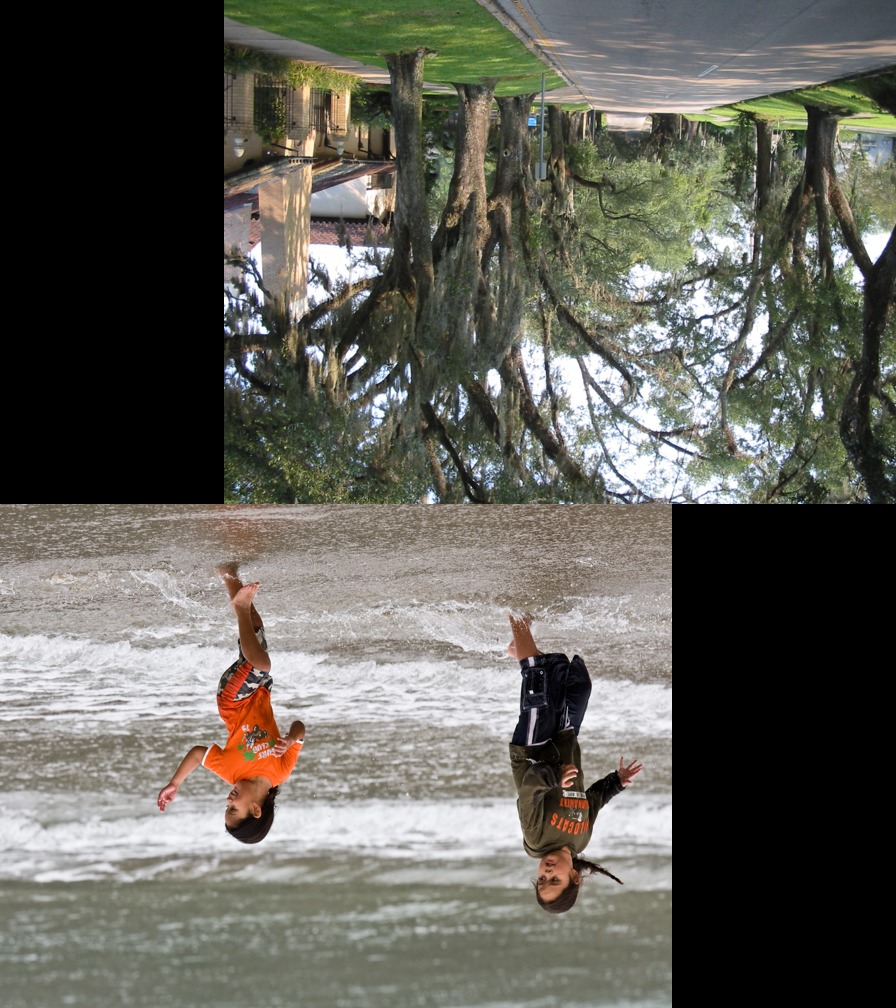}
&
 \includegraphics[width=0.32\columnwidth,height=0.3\columnwidth]{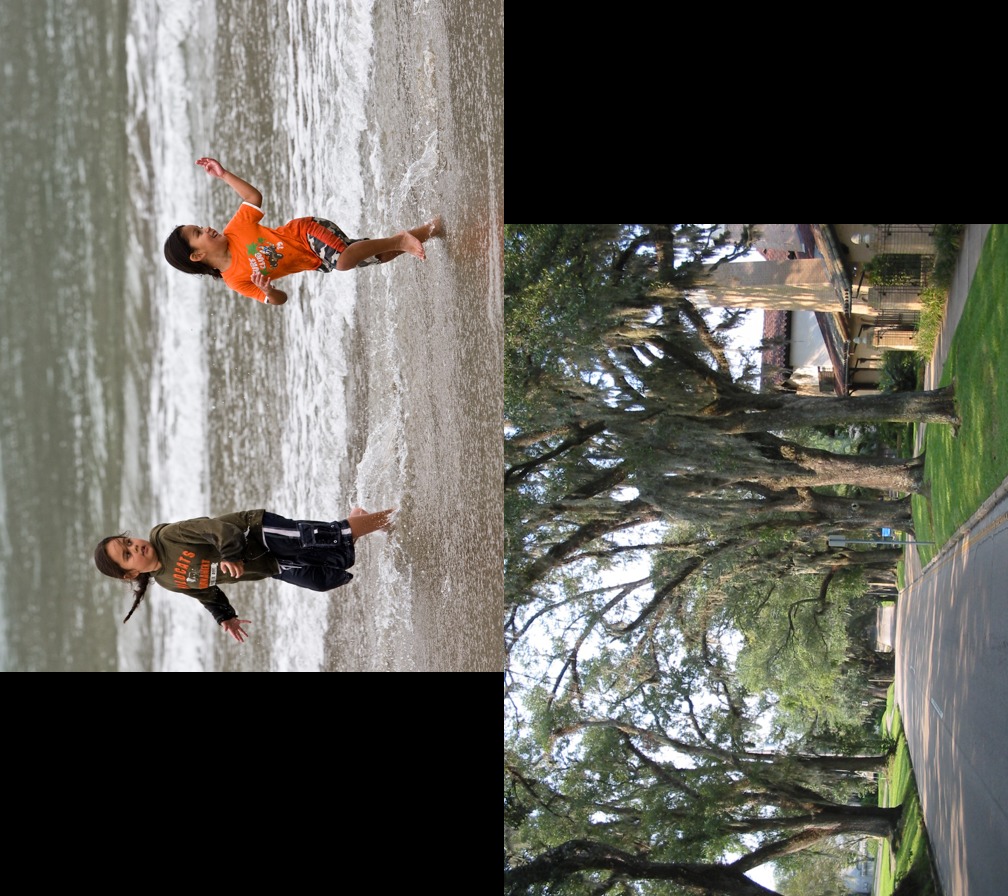}
&
 \includegraphics[width=0.32\columnwidth,height=0.3\columnwidth]{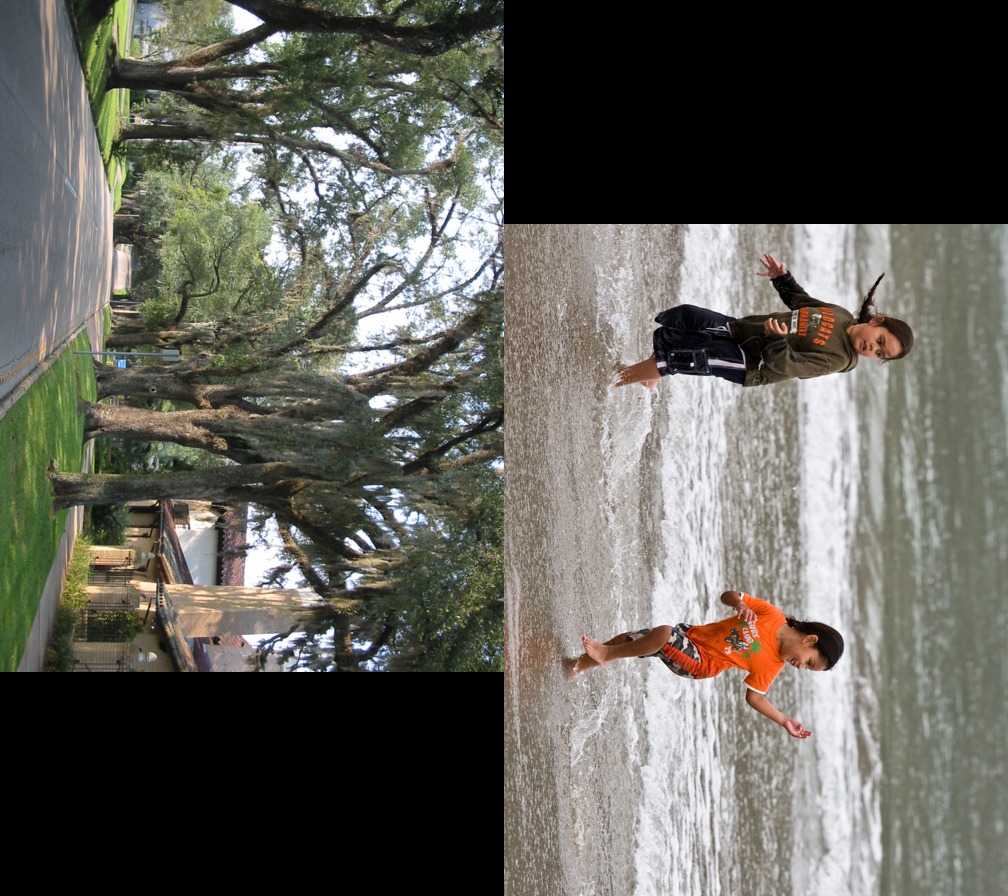}
\end{tabular}
\end{center}
\caption{ \label{fig:mixed} Reconstruction on mixed Type 2 puzzle (864 mixed pieces based on
 two images from MIT dataset.) }
\end{figure*}

\begin{figure*}
\begin{center}
\begin{tabular}{cc}
 \includegraphics[width=1.0\columnwidth]{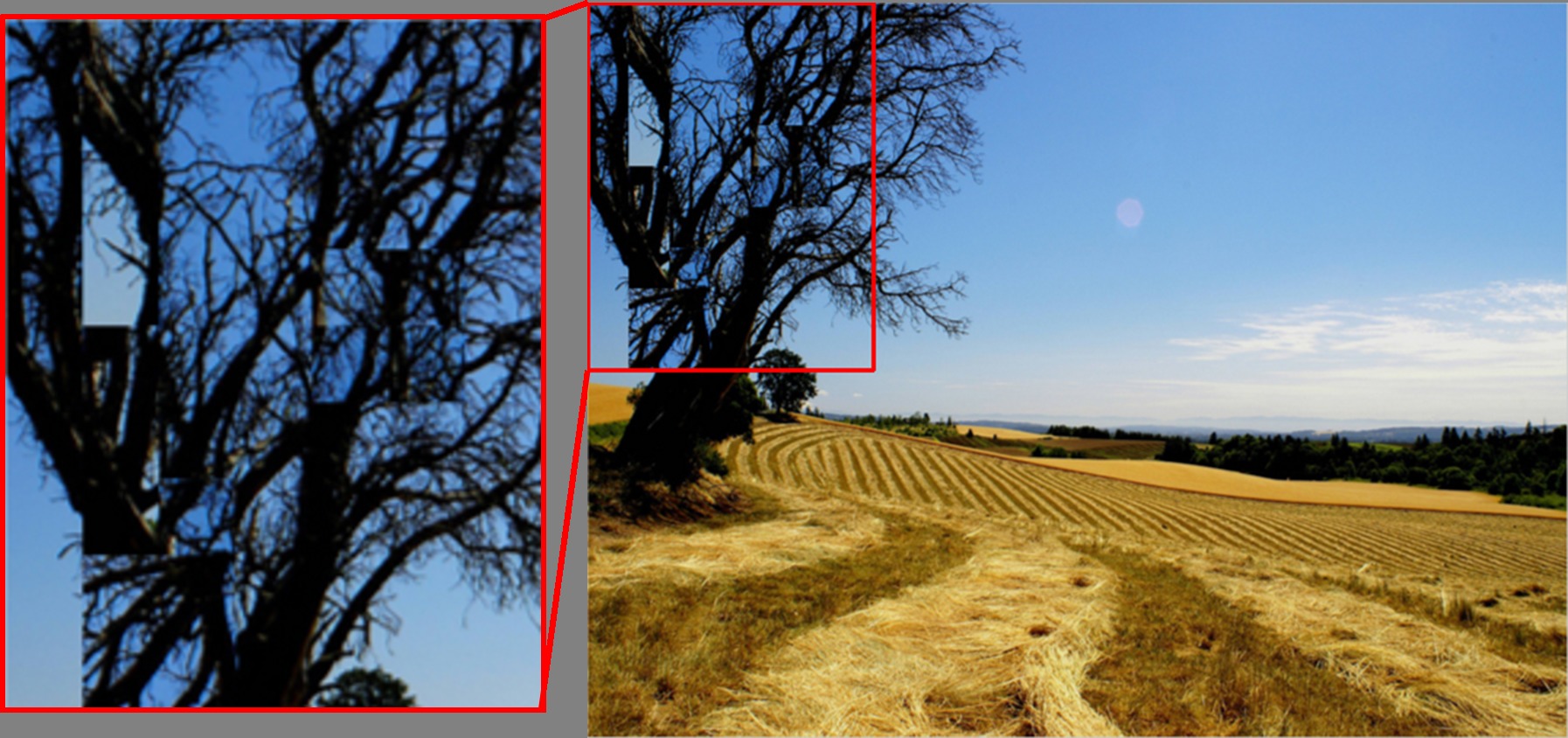}
 &
 \includegraphics[width=1.0\columnwidth]{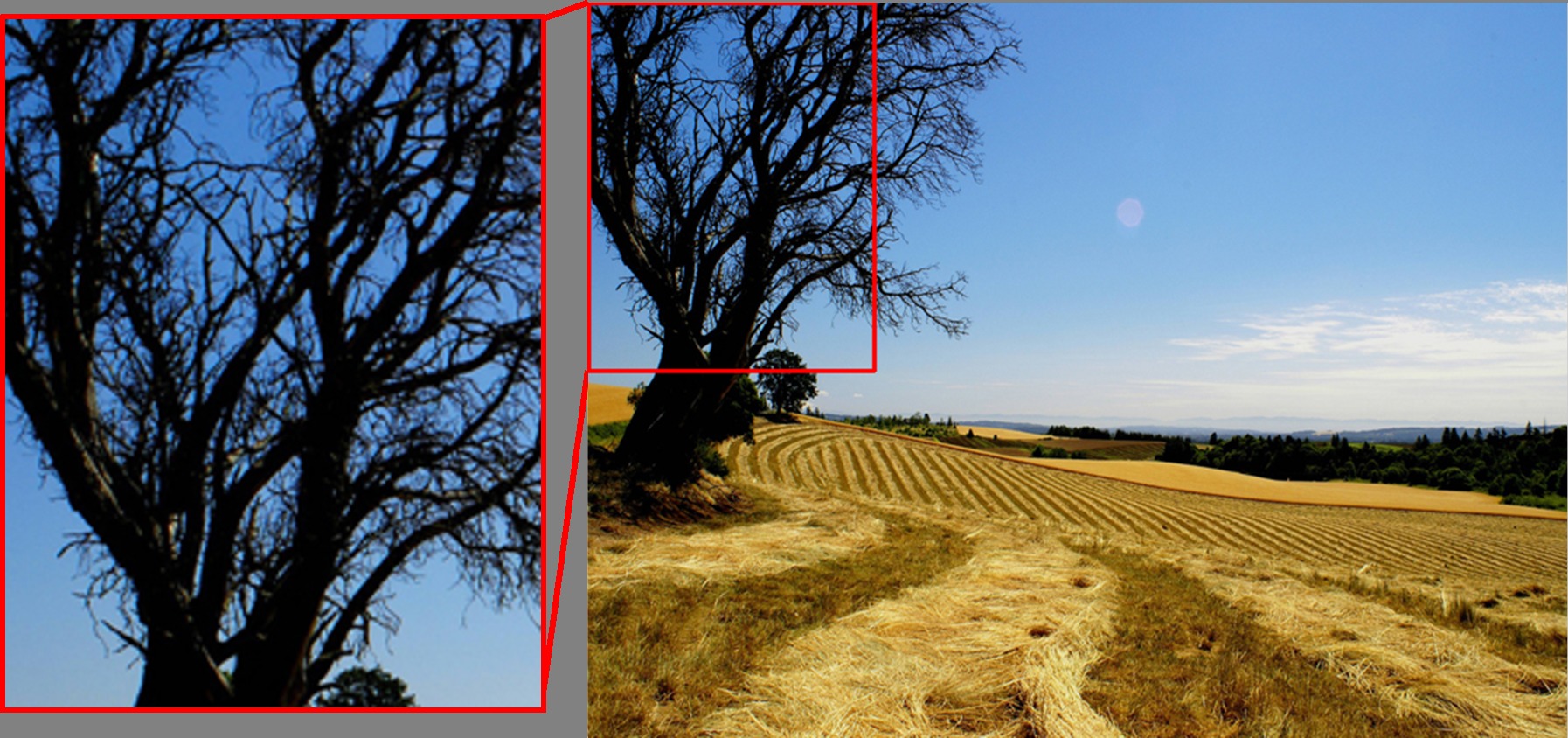}
 \\
 \includegraphics[width=1.0\columnwidth]{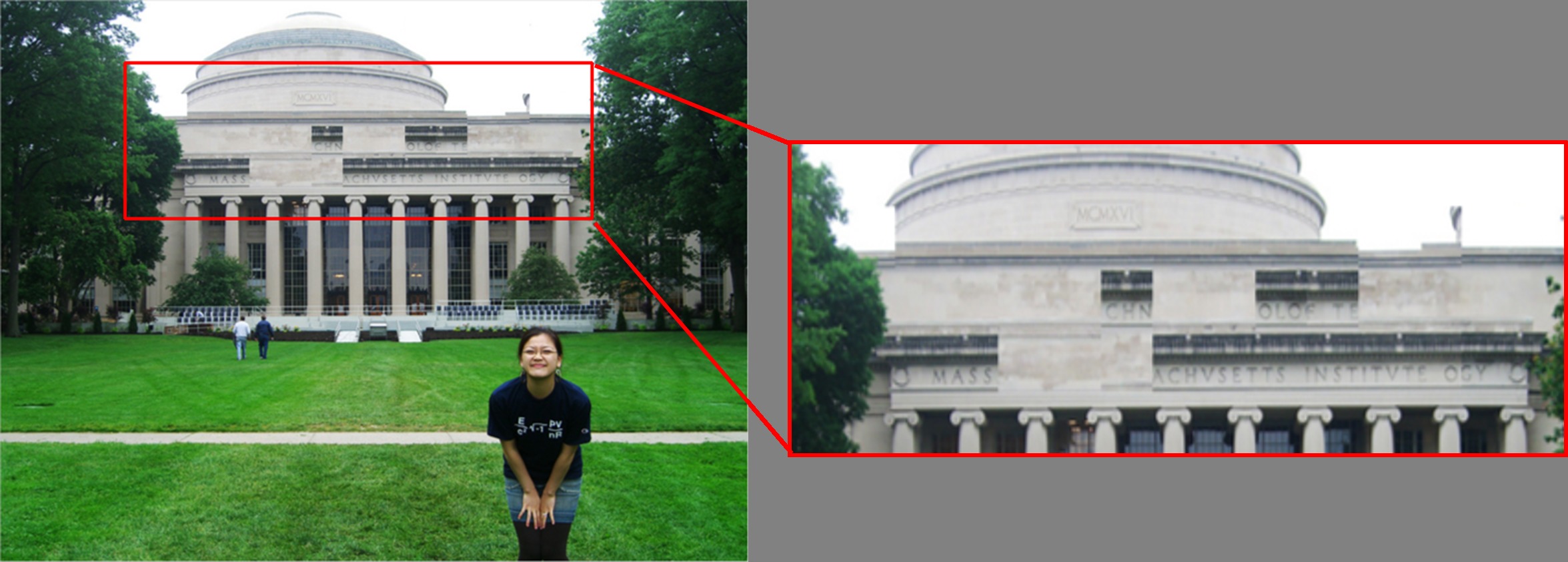}
 &
 \includegraphics[width=1.0\columnwidth]{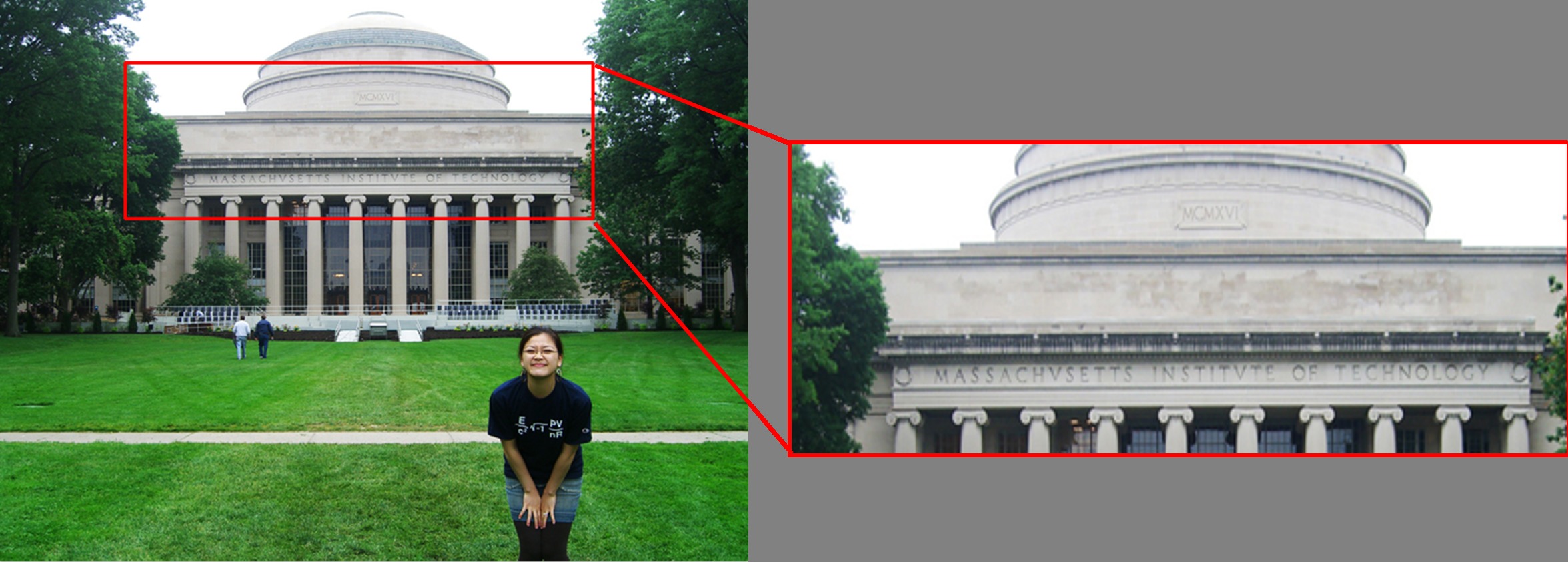}
\\
Son \emph{et al.}~\cite{Son:ECCV2014}  & Ours
\end{tabular}
\end{center}
\caption{ \label{fig:Example} 
Comparison  of Type 2 puzzle reconstruction results between our approach and Son \emph{et al.}~\cite{Son:ECCV2014}. Examples are selected from MIT dataset.}
\end{figure*}


\subsubsection{Type 2 Puzzles}
To solve the more challenging Type 2 puzzles, where both the position
and orientation of the pieces is unknown, we first convert the problem
into a larger Type 1 puzzle by replicating each piece with all 4
possible different rotations and then solving the four mixed Type 1
puzzles simultaneously.

The only difference with the Type 1 puzzle solver is that in this case we have
an additional constraint: identical pieces with different rotations must not be
assigned to the same connected component. Since this constraint is
non-convex it cannot be imposed in the LP directly. In practice, we take the
piece with the largest pairwise matching confidence weight and fix the
positions of its four rotated versions in the LP to predefined
values (in this work we place the 4 copies of the piece at $[\pm 10e{+4},\pm 10e{+4}]$).
This ensures that these copies can not belong to the same connected component.

\section{Experiments}
\label{sec:experiments}
To demonstrate the performance of our method, we apply our algorithm to five
different datasets. The MIT dataset, collected by Cho \emph{et
 al.}~\cite{Cho:2010}, contains 20 images with $432$ pieces, each
with pixel size $28\times28$. The other four datasets, due to Olmos
\emph{et al.}~\cite{Olmos:2003} and Pomeranz \emph{et
 al.}~\cite{Pomeranz:2011}, include two with  20 images ($540$
and $805$ pieces) and two with 3 images ($2360$ and $3300$ pieces).

The metric we use to evaluate our approach follows those of
Gallagher~\cite{Gallagher:CVPR2012} where four measures are
computed: \textbf{``Direct comparison''} measures the percentage of
pieces that are in the correct position, compared with the ground
truth; \textbf{``Neighbor comparison''} measures the percentage of
correct piecewise adjacencies; \textbf{``Largest Component''} is the
fraction of pieces in the largest connected component with correct
pairwise adjacencies, and \textbf{``Perfect Reconstruction''} is a binary
measure of if all the pieces in the puzzle are in the correct
position and with the correct orientation.

\textbf{Type 1 Puzzles}: We evaluate our approach on a wide variety of
existing datasets. Table~\ref{table:Type1OlmosPomeranz} shows the
comparison results of our algorithm with the two state of the art
methods, Gallagher~\cite{Gallagher:CVPR2012} and
Son~\cite{Son:ECCV2014} on the Pomeranz and Olmos datasets. The scores
reported are the average results over all the images in each set. Our
proposed approach outperforms Son~\etal~\cite{Son:ECCV2014} who had
previously shown that their method was state of the art compared with
competing approaches.  We believe that the improvements over previous
methods comes from our LP assembly strategy. Compared with existing
approaches, the global LP formulation and the robustness of weighted
$L_1$ cost allows us to discard mismatches and delay the placing of
ambiguous pieces leading to better performance.

\begin{figure}[!t]
 \begin{center}
 \includegraphics[width=1.2\columnwidth,keepaspectratio]{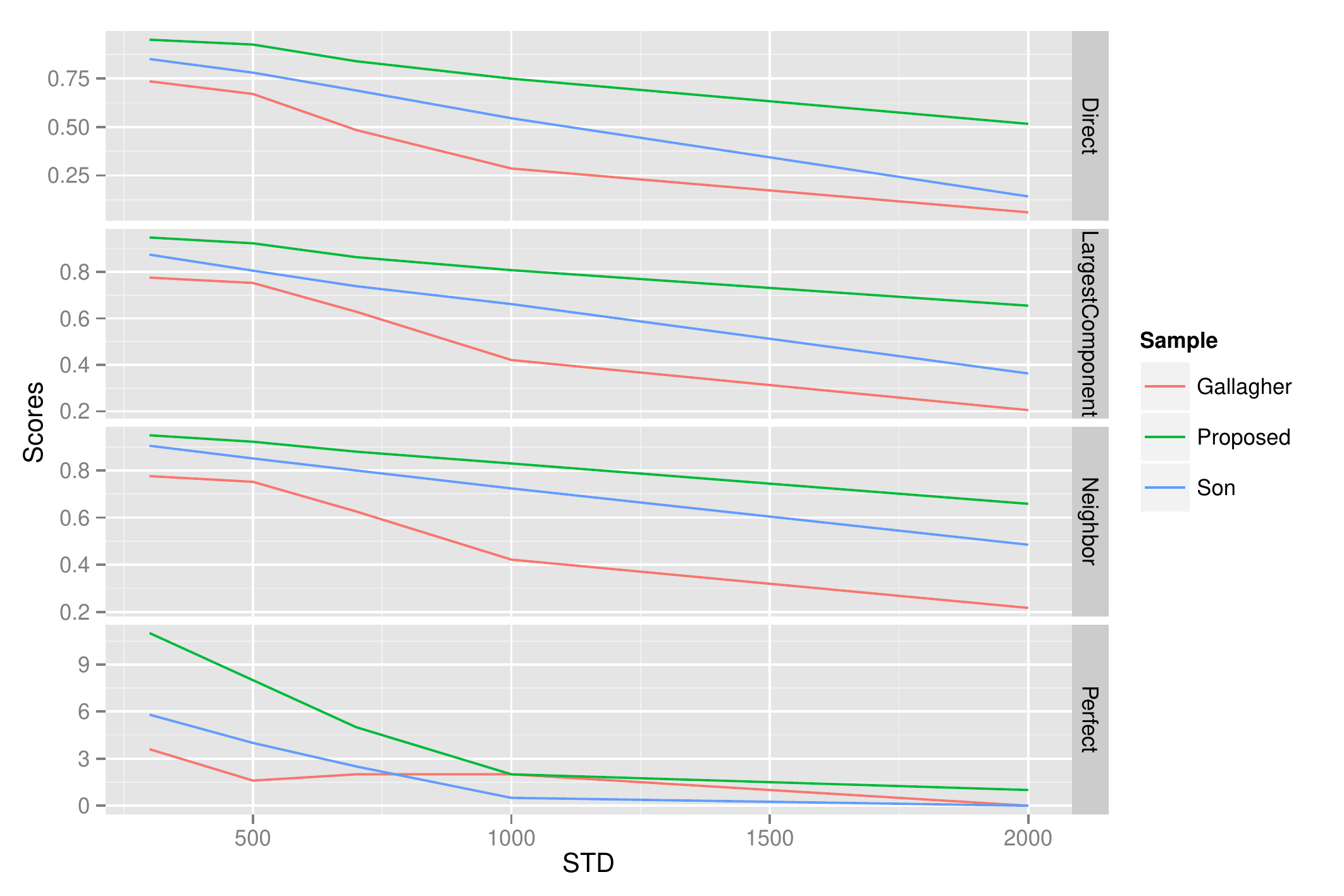}
 \end{center}
 \caption{Comparison of our approach vs
   Gallagher~\cite{Gallagher:CVPR2012} and
   Son~\etal~\cite{Son:ECCV2014} with different amounts of image
   noise.  The X axis shows the standard deviation of the Gaussian
   noise added to the image intensity (intensities range from 0 to
   65535). The Y axis shows the scores on the four performance
   measures. Results are averaged over five runs, and as such the
   number of perfect images may be non-integer. 
 \label{figure:noise}}
\end{figure}

The MIT dataset (Table \ref{table:Type1MIT}) is more saturated but we still outperform all known methods. By adding noise to the images (Figure~\ref{figure:noise}) the advantage of our approach over other greedier methods becomes more apparent.


\begin{table}[t]
\caption{\label{table:Type1MIT} Reconstruction performance on Type 1
  puzzles from the MIT dataset, The number of pieces is K = 432 and
  the size of each piece is P = 28 pixels. The errors reported here
  for Initial LP and Successive LP treat pieces that do not belong to the largest connected component as errors.  As intermediate results need not be
  rectangular, scores are created by sliding the biggest connected
  component across the ground truth image and finding best matching
  locations. }
\begin{center}
\tabcolsep=0.11cm
\begin{tabular}{|c||c|c|c|c|}
\hline
 & Direct & Neighbor & Comp. & Perfect \\ \hline
Cho~\etal~\cite{Cho:2010}&10 \% &55\% & - &0\\ \hline
Yang~\etal~\cite{Yang:2011}&69.2 \% & 86.2 \% & - &- \\ \hline
Pomeranz~\etal~\cite{Pomeranz:2011}& 94 \% & 95 \% & - & -\\ \hline
Andalo~\etal~\cite{Andalo:2012}& 91.7\% & 94.3 \% &-&-\\ \hline
Gallagher~\cite{Gallagher:CVPR2012} & 95.3\% & 95.1\% & 95.3\% & 12 \\ \hline
Sholomon~\etal~\cite{Sholomon:2013} &80.6 \% &95.2\% &- &- \\ \hline
Son~\etal~\cite{Son:ECCV2014} & 95.6\% & 95.5\% & 95.5\% & 13 \\ \hline \hline
Initial LP& 95.1\% & 94.7\% & 95.1\% &13\\ \hline
\footnotesize Successive LPs (constrained) & 95.7\% & 95.5\% & 95.7\% &14\\ \hline \hline
 Constrained              & 95.7\%     & \bf 95.7\%     & 95.7\%     &\bf 14     \\ \hline
 Free         & \bf 95.8\%     & 95.6\%     &\bf 95.8\%     & \bf 14     \\ \hline
\bf Hybrid & 95.7\% & \bf 95.7\% & 95.7\% & \bf 14 \\ \hline \hline
\textcolor[rgb]{1,0,0}{Upper Bound}~\cite{Son:ECCV2014} & \textcolor[rgb]{1,0,0}{96.7\%} & \textcolor[rgb]{1,0,0}{96.4\%} & \textcolor[rgb]{1,0,0}{96.6\%} &
 \textcolor[rgb]{1,0,0}{15} \\ \hline
\end{tabular}
\end{center}
\end{table}
\begin{table}[t]
\caption{\label{table:Type2MIT} Reconstruction performance on Type 2 puzzles from the MIT dataset, The number of pieces is K = 432. Each piece is a 28 pixel square.}
\begin{center}
\begin{tabular}{|c||c|c|c|c|}
\hline
& Direct & Neighbor & Comp. & Perfect \\ \hline
Gallagher ~\cite{Gallagher:CVPR2012} & 82.2\% & 90.4\% & 88.9\% & 9 \\ \hline
Son~\etal~\cite{Son:ECCV2014} & 94.7\% & 94.9\% & 94.6\% & 12 \\ \hline
\hline
Constrained     & \bf 95.6\% &\bf 95.3\% &\bf  95.6\% &\bf 14 \\ \hline
Free & \bf 95.6\% &\bf 95.3\% &\bf  95.6\% &\bf 14 \\ \hline
Hybrid   & \bf 95.6\% &\bf 95.3\% &\bf  95.6\% &\bf 14 \\ \hline
\end{tabular}
\end{center}
\end{table}

Table~\ref{table:Type1MIT}, and Tables 1 and 2 in the suplementary
materials), show the performance of various components of our method.
Looking at Table 1 in the suplementary materials, it can clearly be
seen that our proposed approach, a sequence of LP relaxations,
generates an near perfect solution (all the three metrics are very
close to 100\%) with most of the errors coming from the final filling
step, in which ambiguous pieces must be placed to finish the assembly.

\textbf{Type 2 Puzzles}: As discussed in section $3$, we convert Type 2 puzzles into a mixed
Type 1 puzzles (mixed by four different rotation identical images). Figure~\ref{fig:Type2} shows
a perfect Type 2 puzzle reconstruction from the MIT dataset.
Table~\ref{table:Type2MIT} gives the comparison results of our method with Son~\etal~\cite{Son:ECCV2014} and Gallagher~\cite{Gallagher:CVPR2012} on the
MIT dataset. It can be seen that our approach consistently outperforms both methods. %
 Table~\ref{table:Type2OlmosPomeranz} shows the results on other four additional datasets, where we out-perform Son~\etal~\cite{Son:ECCV2014},
 on all except the last dataset (which only contains three images).

\textbf{Mixed Type 2 Puzzles}: Figure~\ref{fig:mixed} shows the mixed puzzle reconstruction result of our approach.
The input are the $864$ scrambled pieces ($432$ pieces per image) taken from two images of MIT dataset. It can be
seen that our method successfully recovers all the $4$ rotation versions of the two images.

\textbf{Experiment with noise}: 
To further illustrate the robustness of our method, we conduct experiments with image noise on Type 2 puzzles from MIT dataset.
Figure~\ref{figure:noise} shows the comparison of our approach with Gallagher~\cite{Gallagher:CVPR2012} and Son~\etal~\cite{Son:ECCV2014}
at different levels of image noise, where we show  a 
   large gap in performance of our LP-based puzzle solver. This further
   illustrates its clear dominance with respect to competing
   approaches in the presence of noise.

\textbf{Runtime}: 
Our approach takes $5$ seconds for a $432$ piece Type 1 puzzle and $1$
minute for a Type 2 puzzle on a modern PC. The principal bottle-neck
of the algorithm remains Gallagher's $O (n^2)$ computation of pairwise
matches which is common to all state-of-the-art
approaches\cite{Gallagher:CVPR2012,Son:ECCV2014}. As such over the
scale of problems considered, the algorithm complexity scales roughly
quadratically with the number of the pieces of the puzzle.

\section{Conclusion}
We have proposed a novel formulation for solving jigsaw puzzles as a
convergent sequence of linear programs. The solutions to these
programs correspond to a sequence of unambiguous matches that have
been proved to be globally consistent.  Without any significant
increase in runtime our approach outperforms existing
global~\cite{Sholomon:2013,Cho:2010,Yang:2011,Andalo:2012} and greedy
methods~\cite{Gallagher:CVPR2012,Son:ECCV2014} which represent 
state-of-the-art performance. Our system would extend
naturally beyond 2D jigsaws to the reconstruction of damaged 3d
objects\cite{Huang:etal:Siggraph06} and
mosaics\cite{Castaneda:etal:VAST11} and this remains an exciting
direction for future research.  { \bibliographystyle{ieee}
  \bibliography{jigsaw} }
\end{document}